\documentclass[sigconf, anonymous=False]{acmart}

\AtBeginDocument{%
  \providecommand\BibTeX{{%
    \normalfont B\kern-0.5em{\scshape i\kern-0.25em b}\kern-0.8em\TeX}}}

\usepackage{subfigure}
\usepackage{cleveref}
\usepackage[bottom]{footmisc}

\crefname{figure}{Figure}{Figures}
\setcopyright{acmcopyright}
\copyrightyear{2019}
\acmDOI{}
\acmISBN{}
\acmArticle{}
\acmPrice{}

\acmConference[recsysXfashion'19]
{Workshop on Recommender Systems in Fashion, 13th ACM Conference on Recommender Systems}
{September 20, 2019}
{Copenhagen, Denmark}
\acmYear{2019}
\copyrightyear{2019}

\begin{document}

\title{Complementary-Similarity Learning using Quadruplet Network}

\author{Mansi Ranjit Mane}
\email{mansimane5@gmail.com}
\affiliation{%
  \institution{Walmart Labs}
  \streetaddress{P.O. Box 1212}
  \city{Sunnyvale}
  \state{CA}
  \country{USA}
}

\author{Stephen Guo}
\email{sguo@walmartlabs.com}
\affiliation{%
  \institution{Walmart Labs}
  \streetaddress{1 Th{\o}rv{\"a}ld Circle}
  \city{Sunnyvale}
\state{CA}
  \country{USA}}

\author{Kannan Achan}
\email{kachan@walmartlabs.com}
\affiliation{%
  \institution{Walmart Labs}
  \city{Sunnyvale}
  \state{CA}
\country{USA}
}

\renewcommand{\shortauthors}{Mansi Mane, et al.}

\begin{abstract}
  We propose a novel learning framework to answer questions such as "if a user is purchasing a shirt, what other items will (s)he need with the shirt?"  Our framework learns distributed representations for items from available textual data, with the learned representations representing items in a latent space expressing functional complementarity as well similarity. In particular, our framework places functionally similar items close together in the latent space, while also placing complementary items closer than non-complementary items, but farther away than similar items. In this study, we introduce a new dataset of similar, complementary, and negative items derived from the Amazon co-purchase dataset. For evaluation purposes, we focus our approach on clothing and fashion verticals. As per our knowledge, this is the first attempt to learn similar and complementary relationships simultaneously through just textual title metadata. Our framework is applicable across a broad set of items in the product catalog and can generate quality complementary item recommendations at scale. 
\end{abstract}



\maketitle

\section{Introduction}
With growing numbers of online purchases and increases in the variety of items available in e-commerce item catalogs, it has become increasingly necessary for e-commerce companies to offer product recommendations on their websites and online channels. One very important area to item recommendations is the suggestion of complementary items. Whether online or offline, when a customer wishes to purchase a top, (s)he often would be interested in first exploring options for tops, before deciding upon which top to purchase. Once (s)he has purchased the top, additional product suggestions for jeans or jackets can be helpful. Complementary item recommendations serve a variety of purposes, such as reminding customers about other relevant complementary items to purchase, enabling catalog product discovery, and encouraging additional purchases and basket expansion. 

How can we identify if a pair of items are complementary? In practice, functionally complementary items are often purchased together by users. Historical co-purchase data can be used to learn such item-complementary relations; however, there are some challenges \cite{subcomp}. Co-purchase data is usually available for only a small percentage of items in an item catalog, e.g., Pareto principle \cite{marshall201380}. Therefore, modeling approaches primarily relying upon customer interaction data (co-views, co-purchases), such as collaborative filtering, would not properly handle cold-start or low-engagement items.

While learning functional complementary relations, one also needs to learn similarity, to be able to differentiate between complementary items and similar items. If a model is not able to differentiate between the two relations, spurious complementary recommendations can be surfaced. For example, given shoes as an anchor item, one can easily end up recommending different types of other shoes as complementary items, which is not ideal from a customer experience perspective. To address this issue, a common approach is to utilize taxonomy metadata, such as category labels, to differentiate between similar items and complementary items. However, with the increasing number of items and categories available online (more than 600 million on Amazon \cite{amazonstats}), it's difficult to label each and every item with fine-grained category information. This process typically requires a large amount of human labeling from domain experts, along with crowdsourcing budget and support.

Recommendations should be also diverse. e.g. if a customer is shopping for a top, reminding customers about a set of jeans, belt, and scarf is more useful to the user than just recommending a set of complementary jeans. 

We propose a novel Siamese Quadruplet network to address these issues. Our approach learns latent representations where for a given anchor item, similar items are clustered together. Complementary items are clustered together, but are farther than similar items, while non-complementary and non-similar items (we call them negative items) are placed further apart. We use item title information to generate initial representations. 

For this study, we create a dataset of item quadruplets (anchor, similar, complementary, and negative) from the Amazon co-purchase data released by \citet{dyadic}. Using this labeled dataset, we train our proposed Siamese Quadruplet network. Our results indicate that our network is able to differentiate well between similar products and complementary products.

In summary, our contributions are as follows:
\begin{itemize}
    \item We propose a new Siamese Quadruplet framework which is capable of differentiating between similar, complementary, and negative items.
    \item We create a dataset of similar, complementary and negative items to evaluate our approach and compare our proposed approach with various baselines.
\end{itemize}
\section{Literature Review}
\textbf{Compatibility Learning: } Compatibility Learning methods focus on learning compatibility or style; i.e., given a t-shirt and jeans predict if they go well together, while using mostly visual features to learn the style. \citet{dyadic} utilize Siamese loss to learn compatibility relationships. \citet{monomer, typeaware} extend this approach by learning multiple embedding spaces to capture different types of styles. \citet{mcauley2015image} model user preferences as well, while utilizing low rank Mahalanobis transform to learn style space. \citet{shih2018compatibility} describe a generative approach to learn compatibility between items. \citet{bilstm} propose a sequential LSTM based approach to learn compatibility by providing sequences from top to bottom.  \citet{scene_based} utilize images features from detected bounding boxes of  respective items in an image, and then uses attention and siamese-triplet based loss to calculate compatibility between items. For all these approaches, it's assumed that the model has to rank compatibility between items belonging to different categories. 

\textbf{Complementary Learning: }
The objective of these methods is to learn functional complementary relations. 
\citet{quality-aware} utilize a multimodal approach (images, text, user ratings) to improve the quality of recommendations. \citet{triple2vec} learn content un-aware embeddings for users and items to learn both complementary and compatibility relations.

\textbf{Similarity Learning: }
These methods focus on learning similarity between items.
\citet{prod2vec} learn similarity space by using a skip-gram model on item sequences from email receipts.
\citet{linden2003amazon, sarwar2001item, wei2017collaborative} utilize collaborative filtering with different similarity functions like cosine, Pearson similarity etc.
In the computer vision domain, \citet{bell2015learning} propose triplet loss for similarity learning. \citet{quad_cv, son2017multi} describe using quadruplet loss for learning visual similarity. Note that our formulation of quadruplet loss is very different than these approaches. 

\textbf{Complementary-Similarity Learning: }
\citet{subcomp} learn similarity and complementary relations by using latent dirichlet allocation (LDA) \cite{lda} distributions from review and product description data. \citet{linkedvar} improve upon this approach by using linked variational autoencoders, while still relying on reviews data. It is important to note that on many e-commerce websites such rich textual data may not be available, even for training models. Title information is easily available, as compared to product reviews and detailed product descriptions. Our quadruplet network utilizes only widely available title information to learn a shared feature space for both complementary and similarity relationships.

\section{Dataset Generation}
To train a Siamese quadruplet network, we require quadruplets of anchor ($a$), similar ($s$), complementary ($c$), and negative ($n$) items. 
We utilize the Amazon Clothing, Shoes, and Jewelry dataset released by \citet{dyadic} for anchor, complementary, and negative items. The dataset contains item category information, along with titles and images. 
For an anchor item and complementary item pair to be functionally complementary, the items need to be from different categories.  Therefore, we remove the anchor and complementary pairs, which are from same category. We then generate similar items to each anchor item by sampling items from the same category as the anchor. \footnote{The  code and data are available at \href{https://github.com/mansimane/quadnet-comp-sim}{https://github.com/mansimane/quadnet-comp-sim}  
} We have a total of 376,999 unique items and 3,693,416 quadruplets in our dataset. Train and test datasets are generated by splitting quadruplets based on anchor items. The train data has 90\% of the anchor items, while the test data has the remaining 10\% of the anchor items.

The dataset generation proposed here is different than the one used in \citet{subcomp}, where the train-test split is done based on links. Splitting data based on links results in anchor items during training to appear in the test data as well. We make sure that no anchor item in the test data appears in the train data, making the task more realistic and difficult.

\begin{figure}[h]
  \centering
  \includegraphics[width=\linewidth]{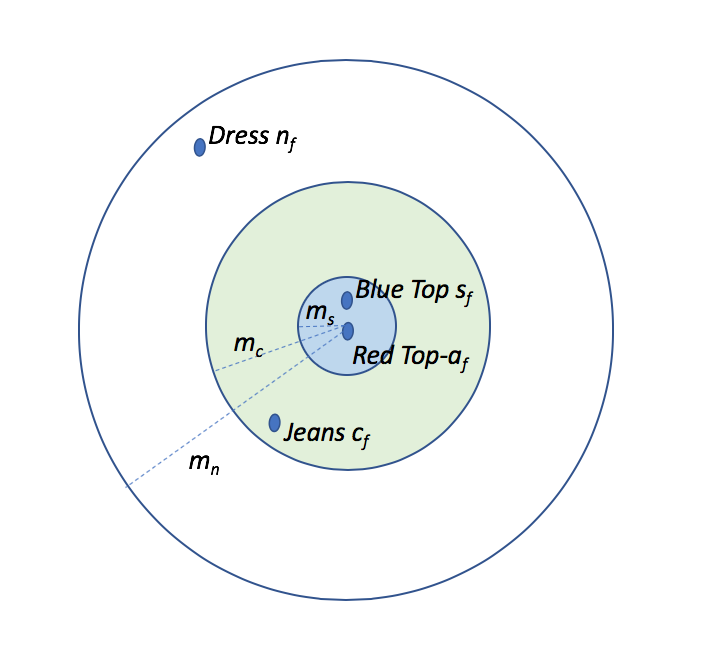}
  \caption{Expected representation space}
  \Description{Expected representation space where similar items are clustered together}
   \label{fig:embed_space}
\end{figure}

\section{Methodology}
We utilize item title information for given $\{a,c,s,n\}$ quadruplets, as text is more helpful to capture functional complementary relations than images and also has good coverage as compared to other attributes as shown in Table ~\ref{tab:cov}.  The text feature vectors $\{a_t,c_t,s_t,n_t\}$  are generated from title information using the Universal Sentence Encoder \cite{guse}. We then learn a mapping function $f$ which projects these text feature vectors into representations which model functional complementariness and similarity between items. Our goal here is to learn representations in a space where we can differentiate between similar, complementary and negative items. Referring to Figure ~\ref{fig:embed_space}, the blue top and red top should be close together as both are functionally similar items. Jeans complement tops, so jeans should also be mapped close to the blue top, but not as close as the red top. On the other hand, dresses are not functionally complementary to tops, so should be placed far away from tops. In this study, we denote the representations learnt through the mapping functions as $\{a_f,c_f,s_f,n_f\}$.

\begin{table}
  \caption{Coverage - Percentage of items for which attribute information is present}
  \label{tab:cov}
  \begin{tabular}{ccl}
    \toprule
    Attribute& Coverage \\
    \midrule
    Item id (asin)  & 100\\
    Image & 99.99\\
    Description & 5.68\\
    Title & 99.95\\
    Price  & 38.29\\
    Brand  & 6.25\\
  \bottomrule
\end{tabular}
\end{table}

\subsection{Siamese Networks}
\begin{table}
  \caption{Notations}
  \label{tab:freq}
  \begin{tabular}{ccl}
    \toprule
    Notation&Description\\
    \midrule
    $a$ &  Anchor item\\
    $c$ & Complementary item to anchor item\\
    $s$ &  Similar item to anchor item\\
    $n$ & Negative item to anchor item\\
    $a_t, c_t, s_t, n_t$ & Text feature vectors for $a,c,s,n$\\
    $a_f, c_f, s_f, n_f$ & Learnt feature representation for $a,c,s,n$\\
    $a_f', c_f', s_f', n_f'$ & Normalized learnt feature representation for $a,c,s,n$\\

  \bottomrule
\end{tabular}
\end{table}

Siamese networks have been utilized in the computer vision community to learn similarity between two images.  \citet{dyadic}  use Siamese triplet networks to learn compatibility between items using visual features. We discuss same framework for learning complementary relation.
 The loss function of a Siamese triplet network minimizes the distance between the anchor (given) item and complementary item, as compared to the distance between the anchor item and negative item. We denote $a_f$ as the embedding projection for the anchor item, $c_f$ as the embedding for the complementary item, and $n_f$ as the embedding for the negative item. The siamese-triplet loss is given as: 

\begin{equation}
    L = max(d(a_f,c_f) - d(a_f,n_f) + margin, 0)
\end{equation}
The margin defines how far away negative items should be from complementary items. This loss is good for similarity learning, however for learning functional complementariness, it has the following challenges:
\begin{itemize}
    \item When $d(a_f,c_f) = 0$, and $d(a_f,n_f)= margin$ or $d(a_f,n_f)> margin$, the loss is zero. This causes complementary items to project onto the anchor item. For example, if the anchor item is blue jeans and complementary item is belt, we do not want their distance in embedding space to be zero, as this will hinder the ability to differentiate between the current anchor jeans and other jeans; i.e., we do not want $d(a_f,c_f) = 0$ as both are different items. 
    
\item Consider the following two cases:  case A - when $d(a_f,c_f)=1.8$ and  $d(a_f,n_f) = 2$ and $margin = 0.2$ and case B-  when $d(a_f,c_f)=0.2$ and  $d(a_f,n_f) = 0.4$ and $margin = 0.2$ . Even though the distance between complementary items is more in case A as compared to case B, the loss would be zero in both cases. The network is not learning anything in case A.
\end{itemize}

We overcome these drawbacks of Siamese triplet loss by incorporating similarity learning, and proposing a novel multitask quadruplet loss for our task.

\subsection{Quadruplet Loss}

Given an item anchor $a$ and it's mapping $a_f$, our goal is to have its similar item $s_f$ not farther than a specific distance, called the margin $m_s$ in the mapped space. This can be achieved by minimizing distance between the anchor item and similar item. We first normalize the embeddings $a_f, c_f, s_f$ and $n_f$ to unit norm, denoted by $a_f', c_f', s_f', n_f'$ and then calculate Euclidean distances between those embeddings. Thus, we write similarity based loss as follows:

\begin{equation}
    L_{sim} =  max(d(a_f',s_f') - m_s, 0) 
\end{equation}

Here we denote Euclidean distance between normalized learnt feature vectors by $d(a_f,s_f)$.

For complementary items, we would like the complementary items mapping $c_f$ closer to anchor item $a_f$, but farther than similar item $s_f$. Thus, we develop the loss between anchor and complementary items as follows.

\begin{equation}
L_{comp} =  max(d(a_f',c_f') - m_c, 0) + max(margin_s - d(a_f',c_f'), 0) 
\end{equation}
Here, the first term in the loss will be positive only when the complementary item is away from the anchor item by more than margin $m_c$ distance, thus making sure the distance between anchor item and complementary item is smaller then $m_c$. The second term will be positive when the distance between anchor item and complementary item is less than margin $m_s$, this prevents complementary items projecting onto similar items.

Finally, we want negative items farther from the anchor item than similar and complementary items. We do not want to penalize the model if negative items are farther than a specific distance, let's say margin $m_n$, otherwise this may hamper learning of similar and complementary items. 

The loss for negative items is modeled as follows
\begin{equation}
L_{neg} =  max(m_n - d(a_f',n_f'), 0) 
\end{equation}
 One of the constraints here in the proposed set up would be $m_s < m_c < m_n$. We also add $l2$ regularization on the weights in the projection matrix. Our projection matrix is a two layer neural network.
The final loss, $L$ is given by:
\begin{equation}
L = L_{sim} + L_{comp} + L_{neg} + \lambda L_{l2}
\end{equation}

\begin{table*}
  \caption{Euclidean distance statistics - between anchor items and their similar, complementary \& negative items}
  \label{tab:dist}
  \begin{tabular}{ccccccccl}
    \toprule
   &\multicolumn{2}{c}{Similar}&\multicolumn{2}{c}{Complementary}&\multicolumn{2}{c}{Negative} &\\
    \midrule
    Data&Mean &Std Dev  &Mean&Std Dev &Mean&Std Dev &cnt\\
   \midrule
    Train Data Before training &    0.82119 & 0.17611   & 0.81975 & 0.15910     & 0.99804 & 0.13286 & 3359252 \\
    Test Data Before training &     0.82752 & 0.17853   & 0.83086 & 0.15937     & 1.0037 & 0.12949 & 334164\\
    
    Train Data After training &     0.24069 & 0.11226   & 0.45845 & 0.11485     & 0.86774 & 0.27724 & 3359252\\
    Test Data After training &      0.24772 & 0.11485   &  0.45181 & 0.09963    & 0.86023 & 0.27182 & 334164\\
  \bottomrule
  \end{tabular}
\end{table*}

\begin{table*}
  \caption{Evaluation}
  \label{tab:acc}
  \begin{tabular}{cccl}
    \toprule
    Method&Ranking Acc&Complementary Acc&Similarity Acc\\
    \midrule
    Universal Sentence Encoder  & 37.68 &  - & -\\
    
    \citet{dyadic}  & 14.92 &  91.05 & 56.45\\
    Our Approach    & \textbf{67.15} & 86.92 & \textbf{68}\\
  \bottomrule
\end{tabular}
\end{table*}

\section{Experiments}

\textbf{Metrics: }
We evaluate our approach using prediction accuracy on complementary and similar items. We also use metric "ranking accuracy". We predict model has ranked items correctly if Euclidean distances between anchor and similar and complementary items are ranked in proper sequence.
i.e.
\begin{equation}
y = \begin{cases}
  1 \text{ if } d(a_f',s_f') <  d(a_f',c_f') <  d(a_f',n_f')\\
 0  \text{ otherwise }
\end{cases}
\end{equation}

\textbf{Implementation: }
We use the Universal Sentence Encoder \cite{guse} to generate 512 dimensional embeddings using item title. The mapping function used is two fully connected layers with ReLU activation. The first layer has 256 hidden units and the second layer has 128 hidden units. We set a learning rate of 0.001 with batch size 512. $m_s$, $m_c$, $m_n$ are set to $0.1, 04, 0.8$ respectively. We train the model for 30 epochs; training time is approximately 5 hours with a TensorFlow implementation on Tesla K80 GPUs.

\textbf{Baselines: }
We compare our results against text embeddings generated by the Universal Sentence Encoder \cite{guse} and the Siamese triplet network utilized by \citet{dyadic}. 

\begin{itemize}
    \item Universal Sentence encoder: These text representations are generated by training a transformer \cite{vaswani2017attention} based encoder on multiple tasks like semantic text classification, fine grained question classification, sentiment analysis, etc. We only report ranking accuracy on this as we do not have specific thresholds on distances to decide similarity-complementary relationships.
    \item Siamese triplet network: We train the Siamese triplet network from \cite{dyadic} to learn only complementary relationships using text (as compared to compatibility using images). For a fair comparison, we use the same layers, learning rate etc. while comparing with our approach.
\end{itemize}

\begin{figure}[h]
  \centering
  \includegraphics[width=\linewidth]{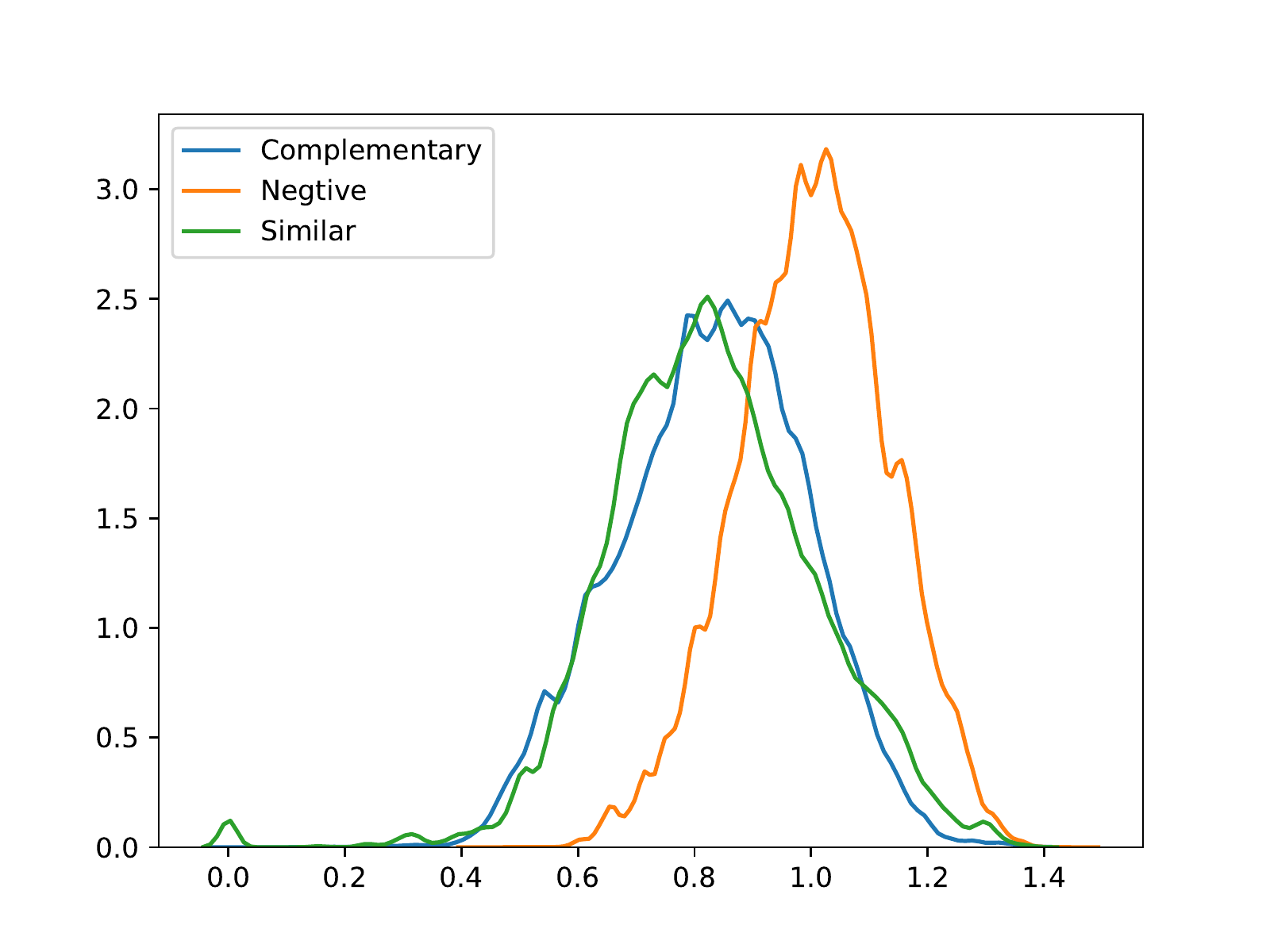}
  \caption{Before training - Euclidean distance distribution between similar, complementary, and negative items }
  \Description{Before training: Euclidean distance distribution between similar, complementary, and negative items. Overlap between similar and complementary items makes it difficult to distinguish between the two}
  \label{fig:euclid_pre_train}
\end{figure}

\begin{figure}[h]
  \centering
  \includegraphics[width=\linewidth]{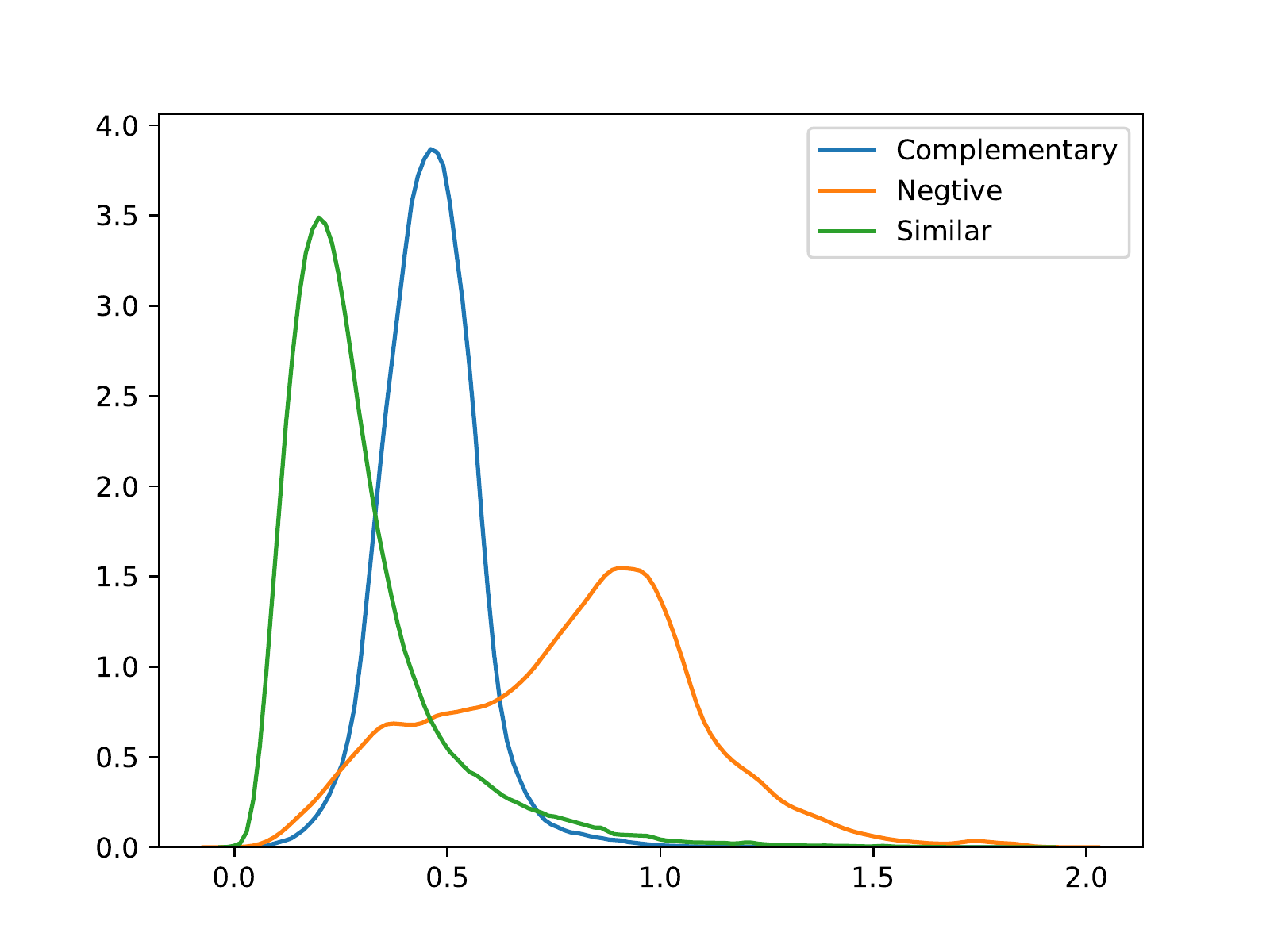}
  \caption{After training - Euclidean distance distribution between similar, complementary and negative items }
  \Description{After training: Euclidean distance distribution between similar, complementary and negative items}
  \label{fig:euclid_post_train}
\end{figure}

  Figure ~\ref{fig:euclid_pre_train} shows the probability density function (pdf) for the Euclidean distance between anchor items and similar items, distance between anchor items and complementary items, and distance between anchor items and negative items for pretrained Universal Sentence Encoder embeddings. Table ~\ref{tab:dist} shows mean and standard deviation for the distance distributions, before and after training. In figure ~\ref{fig:euclid_pre_train} we observe that both similar and complementary items have similar distances, hence it is difficult to differentiate between two. In Figure ~\ref{fig:euclid_post_train}, we show the distance plot for learnt embeddings using quadruplet network. We see that similar and complementary distances are fairly separated after training. The distance distribution between anchor and negative items has more variance. One of the reasons could be that negative items are randomly sampled, which means negative items can also contain some similar and complementary items. We plan to investigate strategic sampling for negative items in our future work.

\begin{figure}
\subfigure{\includegraphics[height=1in,width=1in]{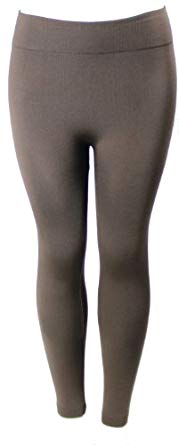}}  \subfigure{\includegraphics[height=1in,width=1in]{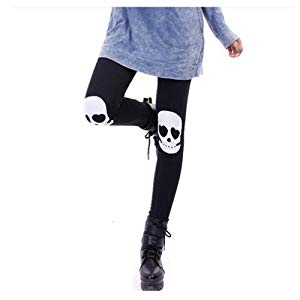}}
\subfigure{\includegraphics[height=1in,width=1in]{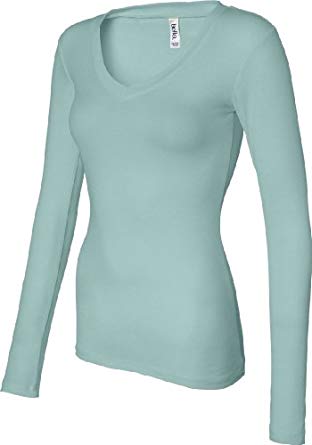}}

\subfigure{\includegraphics[height=1in,width=1in]{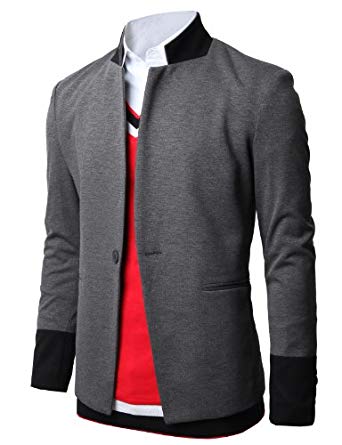}}  \subfigure{\includegraphics[height=1in,width=1in]{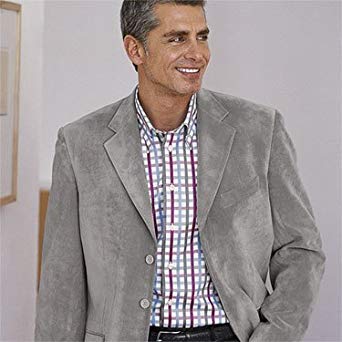}}
\subfigure{\includegraphics[height=1in,width=1in]{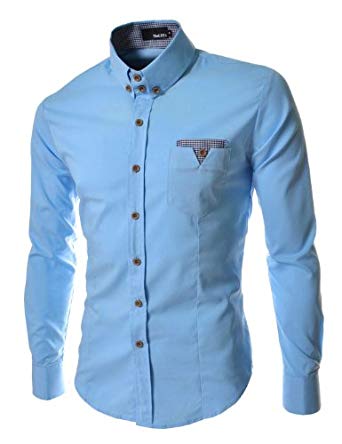}}

\subfigure{\includegraphics[height=1in,width=1in]{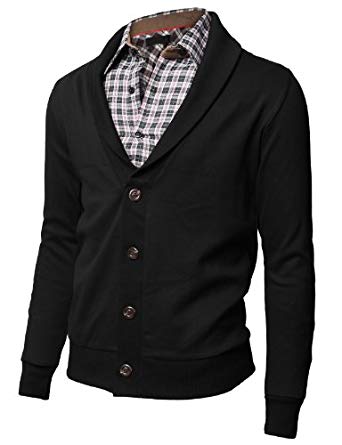}}  \subfigure{\includegraphics[height=1in,width=1in]{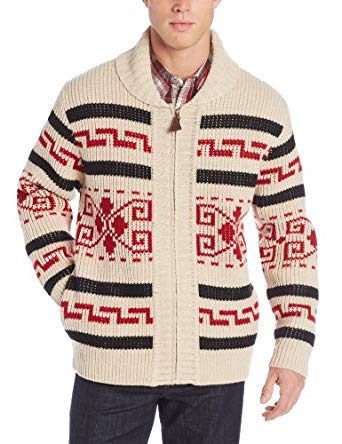}}
\subfigure{\includegraphics[height=1in,width=1in]{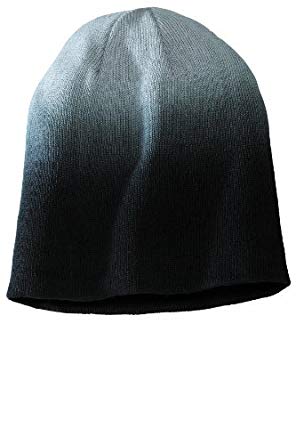}}

\subfigure{\includegraphics[height=1in,width=1in]{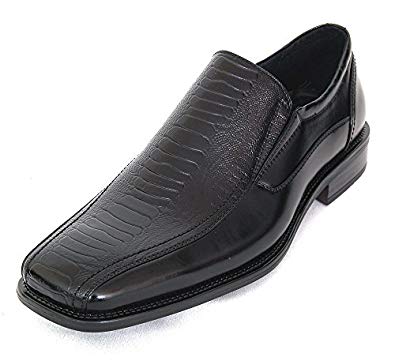}}  \subfigure{\includegraphics[height=1in,width=1in]{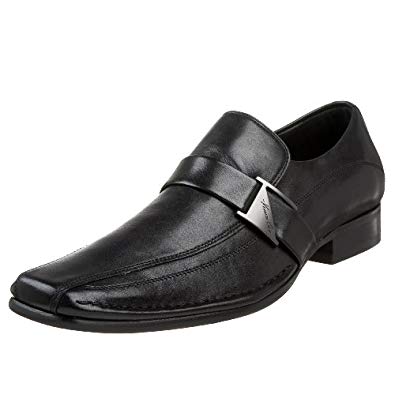}}
\subfigure{\includegraphics[height=1in,width=1in]{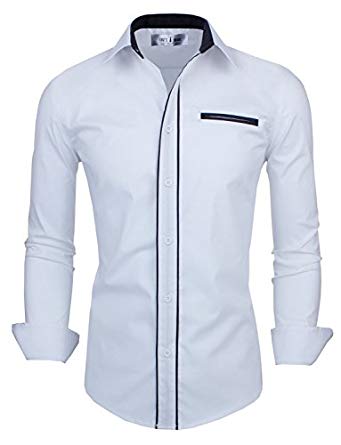}}

\caption{Example model predictions - anchor item (left column), similar item (middle column), and complementary item (right column)}
\label{fig:item_imgs}
\end{figure}

In Table  ~\ref{tab:acc}, we compare ranking, complementary, and negative accuracy between different approaches. We can see that, when Siamese networks are optimized for the complementary task only, similarity accuracy is affected, and vice versa. However with our Quadruplet network, we have a good balance between similar and complementary accuracy and the best ranking accuracy. In Figure ~\ref{fig:item_imgs} we show few example predictions from our model, for different clothing items. 


\section{Conclusion and Future Work}
In this study, we propose a novel Siamese Quadruplet network to learn complementary and similarity relations. We utilize item title text, which is widely available item attribute on most e-commerce websites for the task. Our qualitative and quantitative results show that learning complementary and similarity relations together enable better learning of functional complementary relations. Our approach  enables searching for similar items and complementary items for a given anchor item. Our approach also is applicable for cold-start items (items for which no purchase data is available). 

One area of improvement which we do not address is asymmetry in item relationships. For example, if a customer is shopping for a television stand, (s)he likely already has purchased a television. Although televisions are the most co-purchased items for television stands, televisions are not suitable for recommendations in this scenario. As future work, we plan to expand our approach to model asymmetry and also perform larger scale evaluation on various datasets, including the full Amazon dataset and Polyvore outfits dataset.

\bibliographystyle{ACM-Reference-Format}
\bibliography{sample-base}

\end{document}